\def\year{2022}\relax
\documentclass[letterpaper]{article} 
\usepackage{aaai22}  
\usepackage{times}  
\usepackage{helvet}  
\usepackage{courier}  
\usepackage[hyphens]{url}  
\usepackage{graphicx} 
\usepackage{subfigure}
\usepackage{natbib}  
\usepackage{caption} 
\DeclareCaptionStyle{ruled}{labelfont=normalfont,labelsep=colon,strut=off} 
\usepackage{xcolor}
\usepackage{array} 
\usepackage{booktabs}  
\usepackage{threeparttable}  
\usepackage{multicol}  
\usepackage{multirow}  
\usepackage{natbib}
\usepackage{epsfig}
\usepackage{algorithm}
\usepackage{algorithmic}
\usepackage{amssymb}
\usepackage[utf8]{inputenc}
\usepackage{verbatim}
\usepackage{amsfonts}
\usepackage{amsmath}
\usepackage[switch]{lineno}

\usepackage{multirow}
\usepackage{booktabs}

\usepackage{bm}
\usepackage[utf8]{inputenc}

\usepackage{cleveref}
\crefname{section}{§}{§§}
\Crefname{section}{§}{§§}

\def\ie{\emph{i.e}.~}

\usepackage{newfloat}
\usepackage{listings}
\floatstyle{ruled}
\newfloat{listing}{tb}{lst}{}
\floatname{listing}{Listing}
\nocopyright
\title{Linking-Enhanced Pre-Training for Table Semantic Parsing}
\author {
    Bowen Qin,  \textsuperscript{\rm 1, \rm2 \footnote{Equal contribution.} \footnote{Work done during an internship at Alibaba Group.}}
    Lihan Wang, \textsuperscript{\rm 1, \rm 2 \footnotemark[1] \footnotemark[2]}
    Binyuan Hui, \textsuperscript{\rm 3}
    Ruiying Geng, \textsuperscript{\rm 3} \\
    Zheng Cao, \textsuperscript{\rm 3}
    Min Yang,  \textsuperscript{\rm 1}
    Jian Sun,  \textsuperscript{\rm 3}
    Yongbin Li,  \textsuperscript{\rm 3 \footnote{Corresponding author.}}
}
\affiliations {
    \textsuperscript{\rm 1} Shenzhen Key Laboratory for High Performance Data Mining,\\Shenzhen Institutes of Advanced Technology, Chinese Academy of Sciences \\
    \textsuperscript{\rm 2} University of Chinese Academy of Sciences\\
    \textsuperscript{\rm 3} Alibaba Group\\
    bw.qin@siat.ac.cn, lh.wang1@siat.ac.cn, binyuan.hby@alibaba-inc.com, ruiying.gry@alibaba-inc.com \\
    zhengzhi.cz@alibaba-inc.com, min.yang@siat.ac.cn, jian.sun@alibaba-inc.com, shuide.lyb@alibaba-inc.com
}
\usepackage{bibentry}

\begin{document}
\maketitle

\begin{abstract}
Recently pre-training models have significantly improved the performance of various NLP tasks by leveraging large-scale text corpora to improve the contextual representation ability of the neural network. 
The large pre-training language model has also been applied in the area of table semantic parsing. 
However, existing pre-training approaches have not carefully explored explicit interaction relationships between a question and the corresponding database schema, which is a key ingredient for uncovering their semantic and structural correspondence. 
Furthermore, the question-aware representation learning in the schema grounding context has received less attention in pre-training objective.
To alleviate these issues, this paper designs two novel pre-training objectives to impose the desired inductive bias into the learned representations for table pre-training. 
We further propose a schema-aware curriculum learning approach to mitigate the impact of noise and learn effectively from the pre-training data in an easy-to-hard manner. 
We evaluate our pre-trained framework by fine-tuning it on two benchmarks, Spider and SQUALL. 
The results demonstrate the effectiveness of our pre-training objective and curriculum compared to a variety of baselines.
\end{abstract}

\section{Introduction}
Table semantic parsing~\cite{zelle1996learning,iyer2017learning,dong2018coarse,lin2020bridging} aims to convert natural language questions to SQL, empowering humans to interact with relational databases naturally. It enables people to receive desired information from the large-scale structural database with semantic questions instead of complex code syntax.

Recently, with increasingly complex benchmarks being proposed~\cite{Zhong2017,yu2018spider,yu2019cosql,Yu2019sparc} in the field of table semantic parsing, leveraging the pre-training models~\cite{devlin2018bert} to jointly learn a more powerful representation of natural language question and table schema (including tables and columns) have become a critical problem ~\cite{Yu2020,Shi2020}.

\begin{figure}
    \centering
    \includegraphics[width=8cm]{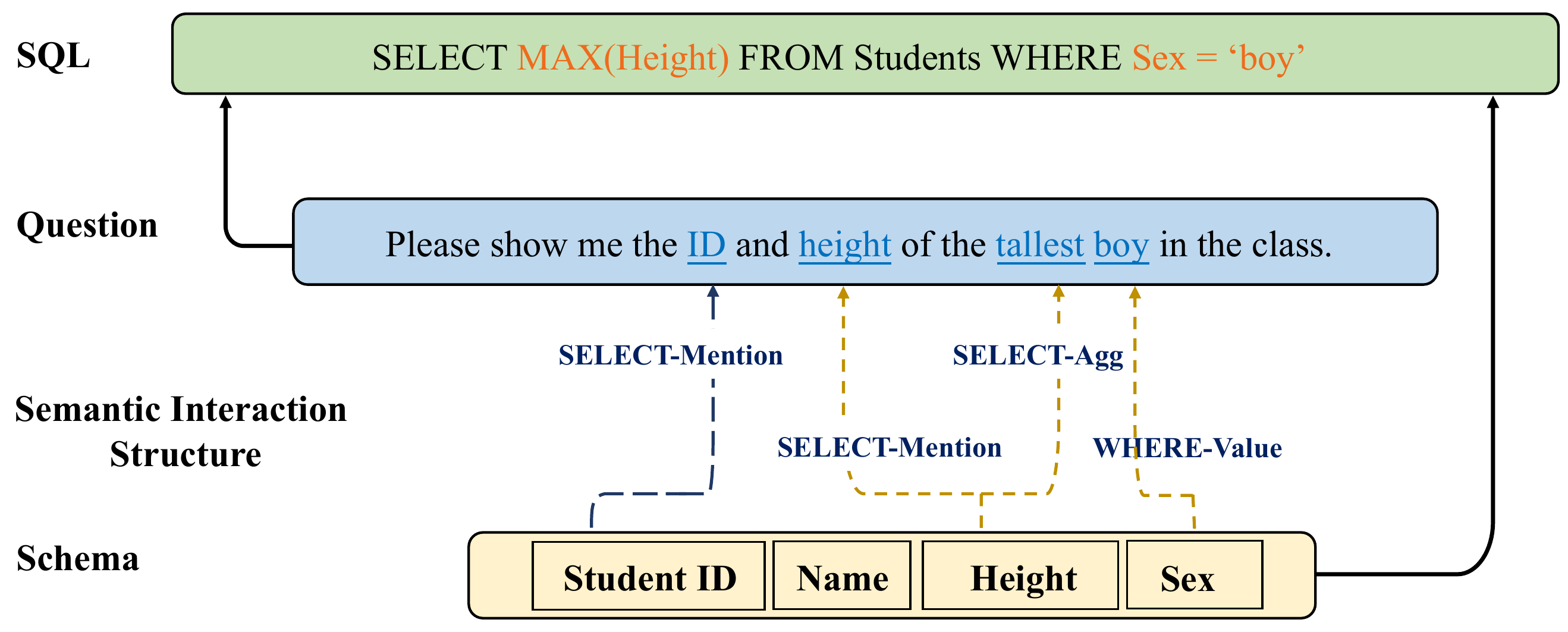}
    \caption{An illustration of how a SQL is generated, and the corresponding semantic interaction structure between question and schema.}
    \label{fig1}
\end{figure}

Existing works ~\cite{Wang2020,lei2020re,chen2019tabfact} on table semantic parsing indicates that the SQL queries bridge natural language questions and the corresponding database schema. As shown in Figure~\ref{fig1}, there exists a semantic interaction structure (schema linking) between question and schema, which have been identified as the current bottlenecks of the parsing task ~\cite{lei2020re}. These method usually solve the schema linking with simple string matching heuristics between question and schema, and integrate it as a sub-component to the complex parsing models, which are not sufficient to grasp more general knowledge.

Recently, some works \citep{Yu2020,Shi2020} proposed a pre-training approach for table semantic parsing to explore the generalization of schema linking.
However, the pre-training objectives mainly predict whether a column or table is used in the question or not, ignoring the explicit SQL-related interaction between the question and schema.
We believe such structure is a crucial ingredient for uncovering semantic and structural correspondences among different components. 
Another prominent challenge is that the modeling of question language understanding in the schema grounding context is not well explored, leading to the difficulty of finding a suitable representation for the question. 

In this paper, we propose a \textbf{S}chema \textbf{D}ependency-enhanced \textbf{CU}rriculum \textbf{P}re-training (named as \textbf{SDCUP}) framework for table semantic parsing, which consists with two novel pre-training objectives, \ie, \textit{Schema Dependency Prediction} ({SDP}) and \textit{Entity Perturbation Recovery} ({EPR}).
Concretely, the SDP mechanism enhances the pre-training process by predicting schema explicit interaction structure between questions and schemas. As shown in Figure~\ref{fig1}, we define the fine-grained dependency edge between \texttt{tallest} and the column \texttt{Height} with label named \texttt{SELET-Agg} through corresponding SQLs and heuristic rules. We then pre-train our model with question and schema input to predict this dependency structure via a biaffine architecture.

Perturbation-recovery self-supervised training paradigm has been shown effective in pre-training \cite{wang2019structbert,lewis2019bart,xuan2021sead}. 
However, previous studies merely consider linguistic aspects for general language understanding. 
Here, we take departure from them by adapting the perturbation-recovery pre-training to table semantic parsing.
We propose a question-aware pre-training task Entity Perturbation Recovery (EPR) to equip the model with the ability of joint reasoning of questions and structural schemas.
Concretely, we identify the question tokens attached to the columns using our predefined directed edges of SDP as the schema-related entities, then we shuffle them and force the model to predict the correct order.

Another thing to be considered is the instance-level noise associated with the data construction and dependency labeling procedures. To tackle the challenge, we propose a \textit{schema-aware curriculum} (SAC) strategy for pre-training, which leverages curriculum learning to alleviate the impact of noise and learns effectively from the pre-training data in an easy-to-hard manner. 

In summary, our contributions are three-fold:

\begin{itemize}
    \item We propose two novel pre-training objectives: Schema Dependency Prediction (SDP) and Entity Perturbation Recovery (EPR). SDP integrates inductive bias for modeling interaction between questions and structured tables into the pre-training process and, EPR provides better generalizability and adaptability question-aware representation. 
    \item We propose a schema-aware curriculum learning mechanism (SAC) at the pre-training stage, leveraging comprehensive valuable information in a noisy corpus.
    \item Extensive experiments on table semantic parsing benchmarks verify the effectiveness of our proposed model. We will make our pre-training data and code publicly available upon the acceptance of this paper.
\end{itemize}

\section{The Proposed Pre-training Framework}

We propose a \textbf{S}chema \textbf{D}ependency-enhanced \textbf{CU}rriculum \textbf{P}re-training (named as \textbf{SDCUP}) framework for table semantic parsing. In the following sections, we will introduce each component of our framework in detail.

\subsection{Construction of Pre-training Data.}

\paragraph{Overall collection.} 
One daunting problem is that the current natural question and corresponding SQL pairs are rare, and only 19.5k real-world data from standard table semantic parsing datasets, Spider~\citep{yu2018spider} and SQUALL ~\citep{shi2020potential} can be used. However, insufficient training data is a central challenge to obtain a powerful pre-training model. Although it is easy to obtain large-scale datasets of the crawled schema (tables from web), it is expensive to obtain SQL queries logically consistent with crawled schema or high-quality questions interrelated with the SQL and schema. 
To tackle these issues, we propose a two-stage sample-and-generate method to generate high-quality question-SQL pairs based on structured schema.
Overall, the pre-training data is comprised of 19.5k real-world data from a standard table semantic parsing dataset includes Spider~\citep{yu2018spider} and SQUALL~\citep{shi2020potential}, and 365k synthetic data includes Spider, WikiTables ~\citep{bhagavatula2015tabel} and synthetic-Spider~\citep{wang-etal-2021-learning-synthesize}.

\begin{figure}[t]
    \centering
    \includegraphics[width=0.5\textwidth]{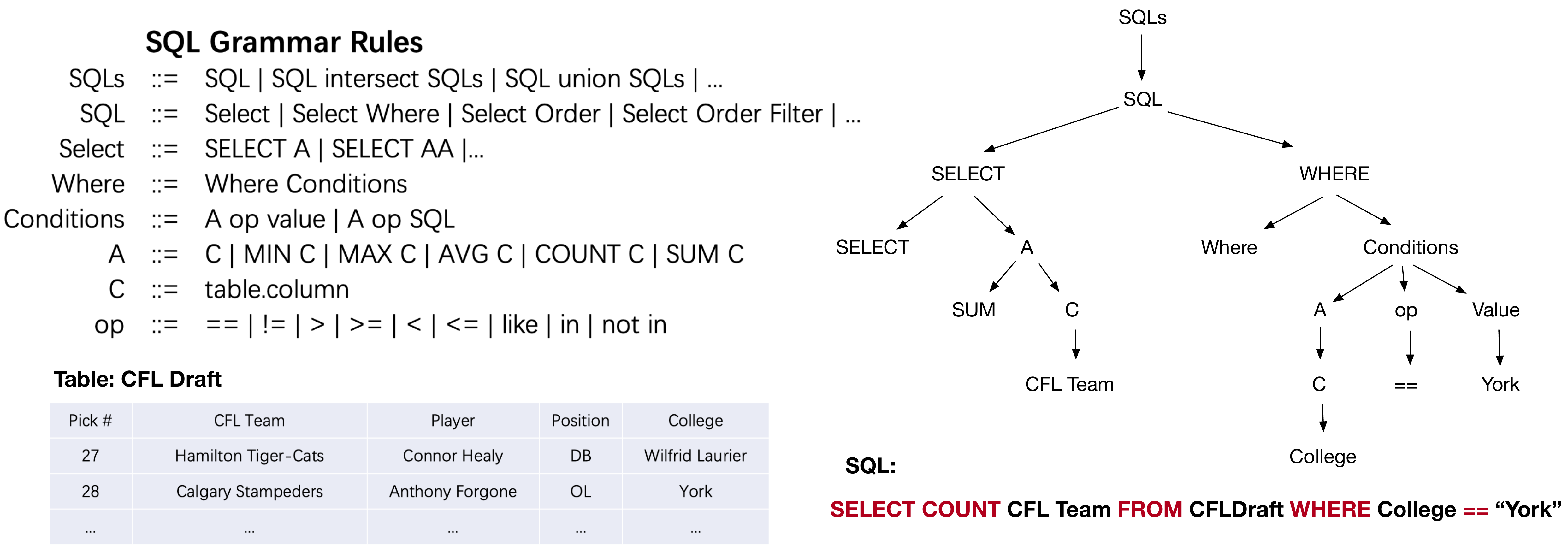}
    \caption{An example of SQL query generation from the grammar. }
    \label{sql2tree}
\end{figure}

\begin{table*}[ht]
    \centering
    \small
    \begin{tabular}{cc}
        \toprule
        Dependency type & Description \\
        \midrule
        \texttt{None} & No dependency. \\
        \midrule
        \texttt{SELECT-Mention} & A question token mentions a column with keyword \texttt{SELECT} without an aggregation function.  \\
        \midrule
        \texttt{SELECT-Agg} & A question token modifies a column with keyword \texttt{SELECT} and an aggregation function.  \\
        \midrule
        \texttt{JOIN-Mention} & A question token mentions a column with keyword \texttt{JOIN}.  \\
        \midrule
        \texttt{WHERE-Mention} & A question token mentions a column with keyword \texttt{WHERE}.  \\
        \midrule
        \texttt{WHERE-Op} & A question token modifies a column with an operation in a \texttt{WHERE} clause.  \\
        \midrule
        \texttt{WHERE-Value} & A question token is a value under a column, and is mentioned in a \texttt{WHERE} clause.  \\
        \midrule
        \texttt{GROUP-BY-Mention} & A question token mentions a column with keyword \texttt{GROUP BY} without an aggregation function.  \\
        \midrule
        \texttt{GROUP-BY-Agg} &  A question token modifies a column with keyword \texttt{GROUP BY} and an aggregation function.  \\
        \midrule
        \texttt{HAVING-Mention} & A question token mentions a column with keyword \texttt{HAVING} without an aggregation function.  \\
        \midrule
        \texttt{HAVING-Agg} & A question token mentions a column with keyword \texttt{HAVING} and an aggregation function.  \\
        \midrule
        \texttt{HAVING-Op} & A question token modifies a column with an operation in a \texttt{HAVING} clause.  \\
        \midrule
        \texttt{HAVING-Value} & A question token is a value under a column, and is mentioned in a \texttt{HAVING} clause.  \\
        \midrule
        \texttt{ORDER-BY-Mention} & A question token mentions a column with keyword \texttt{ORDER BY} without an aggregation function.  \\
        \midrule
        \texttt{ORDER-BY-Agg} & A question token mentions a column with keyword \texttt{ORDER BY} and an aggregation function.  \\
        \midrule
        \texttt{ORDER-BY-Order} & A question token modifies a column with either an ascending or a descending \texttt{ORDER BY}.  \\
        \midrule
        \texttt{LIMIT-Value} & A question token is a value following keyword \texttt{LIMIT}.  \\
        \bottomrule 
    \end{tabular}
    \caption{Dependency types, along with their corresponding descriptions.}
    \label{dep_type}
\end{table*}

\paragraph{Question-SQL pair Synthesize.}
For synthesize process, we first sample SQL queries from schema and then generate natural language questions using a fine-tuned SQL-to-Text BART model~\citep{lewis2020bart}.
An arbitrary natural language question $Q=\{q_i\}$ contains $|Q|$ tokens. A schema $S=\{c_j\}$ is underpinned by $|S|$ columns and each column $c_j$ can be further decomposed to $|c_j|$ tokens. 
We firstly utilize production rules from the SQL grammar to automatically generate SQL queries inspired by \citep{Zhong2017, wang2020chitesql}. As illustrated in Figure~\ref{sql2tree}, the SQL query can be represented as a tree using the rule sequence of \texttt{SQLs = SQL}, \texttt{SQL = Select Where}, \texttt{Select = SELECT A}, \texttt{Where = WHERE Conditions}, and so on, all of which are production rules of the SQL grammar. By exploiting every rule of the grammar, we can generate SQL queries covering patterns of different complexity. 
It is noteworthy that schema in WikiTables are always single-table ones. Hence, before the sampling stage, we select semantically related\footnote{If two tables have shared column names, then we regard them as semantically related.} schema to form multi-table schema for WikiTables. At the generation stage, we adopt a BART model and fine-tune it on SQL-question pairs from Spider. The input is the original SQL and it is directly tokenized by the BART tokenizer without additional pre-processing. After fine-tuning BART, the model can generate high-quality question consistent with the input SQL, achieving a 0.251 BLEU score on the development set. Then we use the model to generate questions for sampled SQLs.m

\paragraph{Schema Dependency Generation.}
We define a {schema dependency} as relational bipartite graphs $\mathcal{G}_\mathcal{S}= \langle \mathcal{V}_\mathcal{S},\mathcal{E}_\mathcal{S} \rangle$, where $\mathcal{V}_\mathcal{S} = Q \cup S$ and $\mathcal{E}_\mathcal{S}$ are schema dependency edges and labels among the question tokens and schema. 
To perform the fine-gained dependency discovery, as shown in Table \ref{dep_type}, we pre-define 17 dependency types based on commonly used keywords, aggregation functions, and operations in SQLs. 
The concrete schema dependency is built with rules (n-gram Levenshtein Distance matching) and well-design trigger function guided by the SQL query. 
For example, given a question ``\textit{show height of the student who is the highest in the class}'' and a schema, a corresponding SQL should be ``\texttt{SELECT MAX(height) FROM student}''. 
Guided by elements mentioned in the SQL, we extract mentioned columns (i.e., \texttt{height}) and find why they are mentioned (i.e., \texttt{SELECT} and \texttt{MAX}), and get corresponding token spans in the question (i.e., \textit{the highest}) that are logically related to the column.
By virtue of these fine-gained types, we can perceived which column should be mentioned and why conduct it easily.

\begin{figure*}[ht]
    \small
    \begin{center}
    \includegraphics[width=0.85\textwidth]{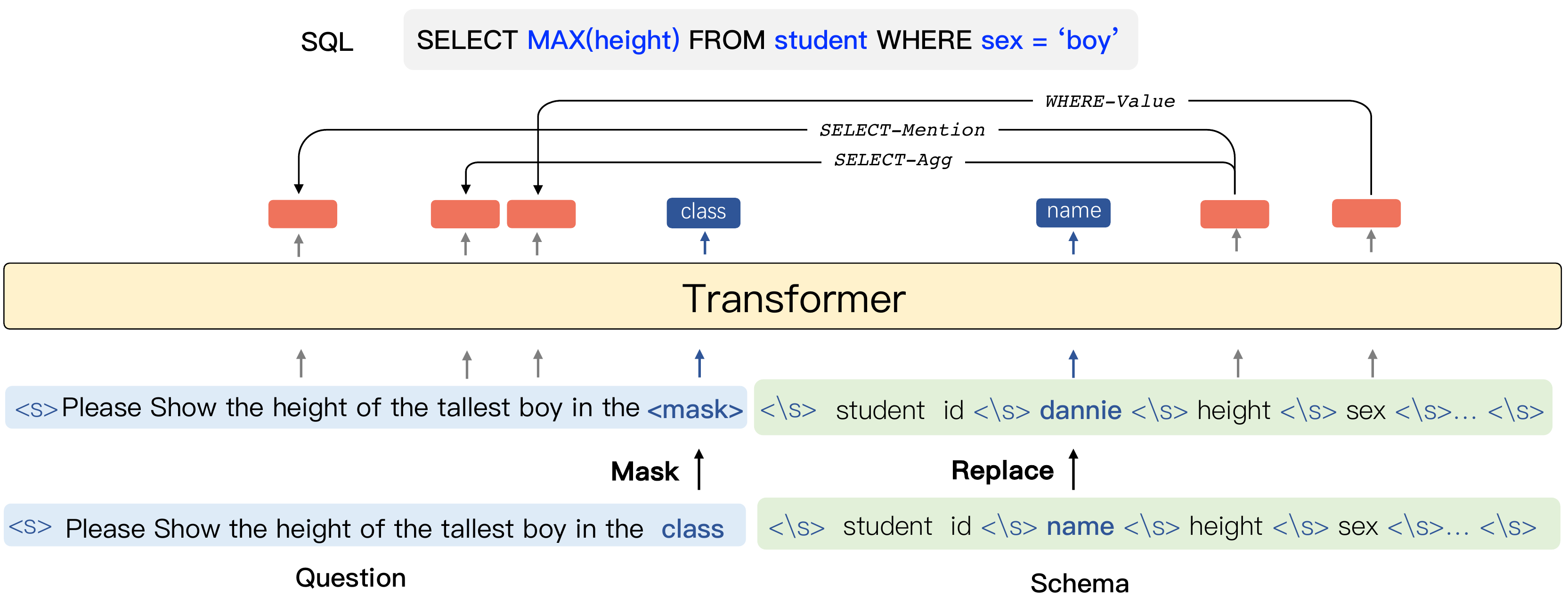}
    \caption{The MLM objective predicts the randomly masked question tokens and column names. The SDP objective decides whether or not a  dependency edge starting from a column and pointing to a question token exists and predicts the relation label for each potential edge. }
    \label{sdcup_1}
    \end{center}
    
\end{figure*}

\subsection{Pre-training Objectives}
\label{SDP}

\begin{figure*}
    \small
    \begin{center}
    \includegraphics[width=0.8\textwidth]{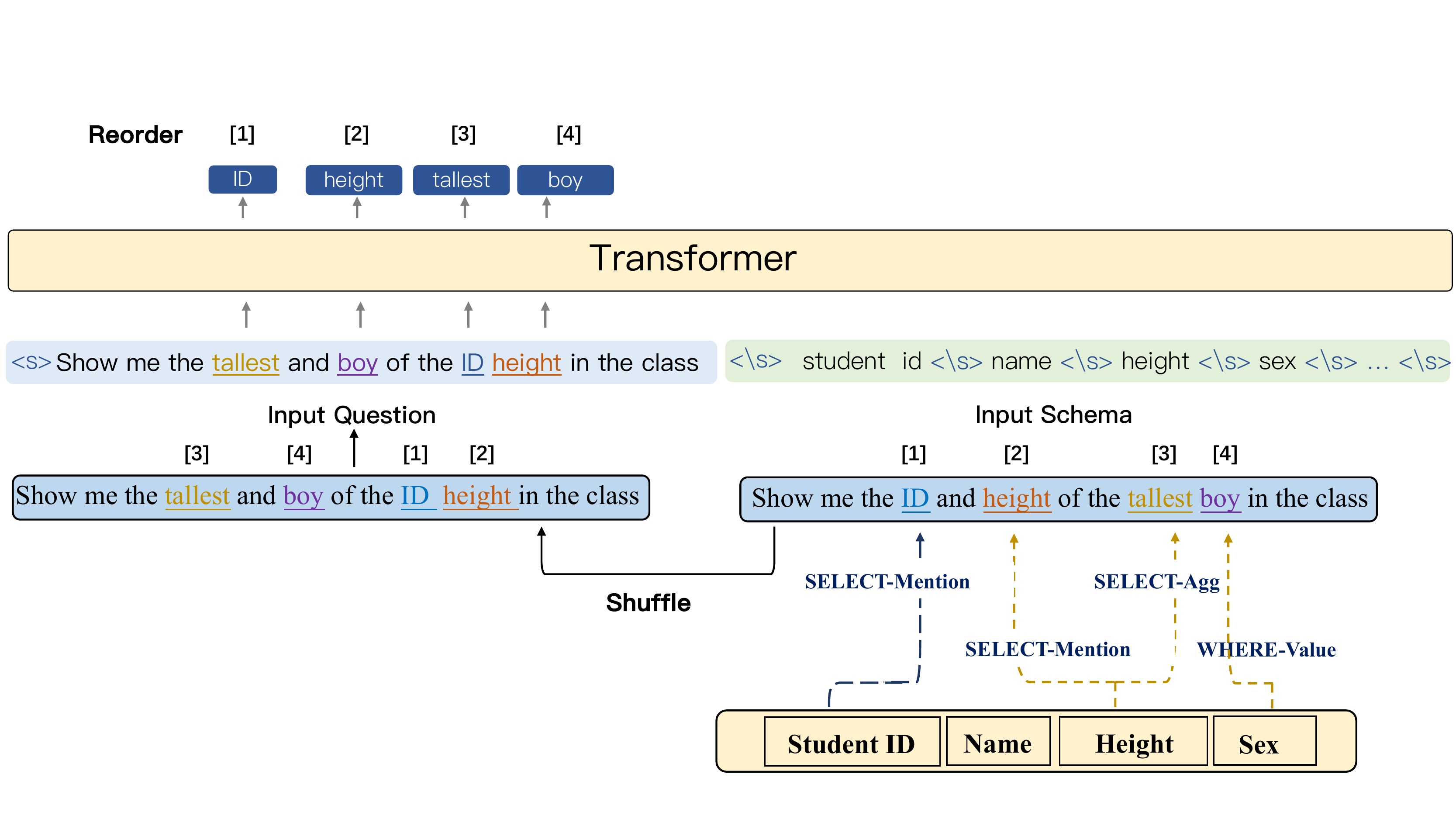}
    \caption{The EPR objective reconstructs the shuffled query sequence with schema-related entities in correct order.}
    \label{sdcup_2}
    \end{center}
    
\end{figure*}

The input fed $I$ into the pre-training model is a sequential concatenation of above-mentioned elements:
\begin{equation}
    I=[\langle\mathsf{s}\rangle;q_1,\dots;\langle\backslash\mathsf{s}\rangle; c_1;\langle\backslash\mathsf{s}\rangle;\dots ;\langle\backslash\mathsf{s}\rangle],
\end{equation}
where $\langle\mathsf{s}\rangle$ and $\langle\backslash\mathsf{s}\rangle$ are special tokens respectively used to denote start and separation of the input.
we first obtain the contextualized representation $\mathbf{H}=\{\mathbf{h}_t\}$ using the last layer output of the RoBERTa \citep{Liu2019} encoder. Representation for each question token $\mathbf{Q}=\{\mathbf{q}_i\}$ is exactly the hidden representation, and representation for each column $\mathbf{S}=\{\mathbf{c}_j\}$ is the hidden representation of the separation token (\ie, $\langle\backslash\mathsf{s}\rangle$) right behind the column's tokens. 

\paragraph{Masked Language Model with Value Replacement.} 
Deriving from the standard MLM task, our Masked Language Model with value replacement objective (\textbf{MLM}) not only preserves the advantage of  maintaining and discovering in-task contextual information as illustrated in \cite{gururangan2020don}, but improves the model’s capability to exploit the implicit semantic relations between the column names and corresponding table values.  
Specifically, the process of masking question token follows the standard MLM procedure by randomly masking $25\%$ of question tokens in each input sequence. Tokens in column names are randomly replaced with tokens drawn from their \textbf{values} also with the $0.25$ probability \cite{Shi2020}. For example, as shown in the Figure \ref{sdcup_1}, the model needs to recover the column name \textit{name} from the corresponding table value \textit{dannie} to capture the interrelationships between value and column. This requires the model to perform deductive logic reasoning and incorporate more domain knowledge.

\paragraph{Schema Dependency Prediction.} 
We design the schema dependency prediction objective (\textbf{SDP}) to explore semantic dependency structures between the questions and schemas. 
As shown in the upper part of Figure~\ref{sdcup_1}, the directed edges and labels have been generated in the data construction stage, in which a pre-defined dependency edge/label starts from a \textit{head} (column) and points to a \textit{dep} (question token).

We use single-layer feed-forward networks (FFN) to compute the representation of ${c}_{j}$ as an edge's head and edge's label head. As such, we can model edge connections and edge labels in detail. We perform a similar operation for ${q}_{i}$.

\begin{equation}\label{mlp}
\resizebox{.6\hsize}{!}{$
\begin{array}{l}
\mathbf{c}_{j}^{(\text {edge-head})} =\operatorname{FFN}_{\text {edge-head}}\left(\mathbf{c}_{j}\right), \\
\mathbf{c}_{j}^{(\text {label-head})} =\operatorname{FFN}_{\text {label-head}}\left(\mathbf{c}_{j}\right), \\
\mathbf{q}_{i}^{(\text {edge-dep})} =\operatorname{FFN}_{\text {edge-dep}}\left(\mathbf{q}_{i}\right), \\
\mathbf{q}_{i}^{(\text {label-dep})} =\operatorname{FFN}_{\text {label-dep}}\left(\mathbf{q}_{i}\right).
\end{array}
$}
\end{equation}

Next, we perform the biaffine attention mechanism to capture the complex dependency between natural language question token $q_i$ and column $c_j$. 
\begin{equation}\label{biaff_forward}
\resizebox{.8\hsize}{!}{$
\begin{array}{l}
\mathbf{s}_{i, j}^{(\text {edge})}=\operatorname{Biaff}_{\text {edge}}\left(\mathbf{q}_{i}^{(\text {edge-dep})}, \mathbf{c}_{j}^{(\text {edge-head})}\right), \\
\mathbf{s}_{i, j}^{(\text {label})}=\operatorname{Biaff}_{\text {label}}\left(\mathbf{q}_{i}^{(\text {label-dep})}, \mathbf{c}_{j}^{(\text {label-head})}\right), \\
\mathbf{y}_{i, j}^{\prime(\text{edge})}=\left\{\mathbf{s}_{i, j}^{(\text{edge})} \geq 0\right\}, \\
\mathbf{y}_{i, j}^{\prime(\text{label})}=\operatorname{softmax} (\mathbf{s}_{i, j}^{(\text {label})}),
\end{array}
$}
\end{equation}
where $\mathbf{y}_{i, j}^{\prime(\text{edge})}$ predicts whether or not a directed edge $q_i \leftarrow c_j$ exists, and $\mathbf{y}_{i, j}^{\prime(\text{label})}$ predicts the best label for each potential edge.
$\operatorname{Biaff(\cdot)}$ is a biaffine function shown in Eq.~\ref{biaff} ~\cite{Dozat2017}, which is a generalization of the bilinear mapping that include also multiplicative interactions between question tokens and schema columns.
\begin{equation}\label{biaff}
\resizebox{.85\hsize}{!}{$
\operatorname{Biaff}\left(\mathbf{x}_{1}, \mathbf{x}_{2}\right)=\mathbf{x}_{1}^{\top} \mathbf{Ux}_{2}+\mathbf{W}\left(\mathbf{x}_{1} \oplus \mathbf{x}_{2}\right)+\mathbf{b}
$,}
\end{equation}
where $\mathbf{U}, \mathbf{W}$, and $\mathbf{b}$ are learnable parameters. The SDP loss $\mathcal{L}^{\mathrm{SDP}}$ is computed as cross-entropy, which contains both the edge prediction loss and label prediction loss.

\paragraph{Entity Perturbation Recovery.} 

Furthermore, we propose a question-aware entity perturbation recovery objective (\textbf{EPR}) as an auxiliary task to recover the input from well-defined perturbation noises. The practical perturbation-recovery approach is carried out during the training process in a self-supervision mode that trains the model to reconstruct the correct sequence given the shuffled one.

As shown in Figure {\ref{sdcup_2}}, pre-defined directed edges are generated from the data construction stage to explain the interrelationships between question tokens and related schema columns. We regard the question tokens attached to the columns by the predefined directed edges of SDP as the \textbf{primary entities} and perturbation them in random order.  
Given the input sequence $I=[\mathbf{Q}, \mathbf{S}]$ shown in Figure {\ref{sdcup_2}}, the perturbation noise randomly re-permutes the schema-related entities in the source question $\mathbf{Q}$ while the schema sequence $\mathbf{S}$ remains unchanged. The perturbation-recovery self-supervised training paradigm of recovering the shuffled primary entities orders forces the model to capture the inner relation between different entities and further improves the modeling of question language understanding in the schema grounding context. The EPR loss $\mathcal{L}^{\mathrm{EPR}}$ is computed as standard cross-entropy loss between the predicted entity order and the ground-truth one.

\paragraph{Overall Pre-training Objective.}
Our final objective joins the standard MLM loss $\mathcal{L}^{\mathrm{MLM}}$ , SDP loss $\mathcal{L}^{\mathrm{SDP}}$ , EPR loss $\mathcal{L}^{\mathrm{EPR}}$  and using a dynamic trade-off approach~\cite{kendall2018multi} to balance these two sub-goals:
\begin{equation}
\resizebox{.85\hsize}{!}{$
\begin{aligned}
    \mathcal{L}=\frac{1}{2\alpha^2}\mathcal{L}^{\mathrm{MLM}}+\frac{1}{2\beta^2}\mathcal{L}^{\mathrm{SDP}}+ \frac{1}{2\gamma^2}\mathcal{L}^{\mathrm{EPR}}+\mathrm{log} \alpha \beta \gamma,
\end{aligned}$}
\end{equation}
where $\alpha$ ,$\beta$ and $\gamma$ are trainable parameters.

\subsection{Schema-aware Curriculum Learning}
\label{sac}
The instance-level noise caused by data construction and automatic dependency labeling impairs convergence stability and performance of pre-training.
Curriculum learning~\citep{bengio2009curriculum} is known to alleviate the effect of noise, which is gradually transitioning to learning more complex concepts based on some pre-defined learning strategy.
Despite the success of curriculum learning on various tasks ~\citep{Platanios2019,wang2020comprehensive}, extending them to semantic parsing tasks is less clear.

Specifically, we propose a schema-aware curriculum (\textbf{SAC}) approach for pre-training. We first calculate the difficulty $d$ of each data instance with $d = |L|$, in which $|L|$ is the length of the input consists of both the question and schema normalized by the Min-Max scaling method \citep{patro2015normalization}.
Note that we use the length of the input instead of the length of the question only because the quality of dependency labeling is highly correlated with both the length of the question and schema. 
Accordingly, we employ the model competence $c$  \citep{Platanios2019}:
\begin{equation}
    \resizebox{.6\hsize}{!}{$
    c(t)=\sqrt{\frac{t(1-\mathrm{min}^2(d))}{T}+\mathrm{min}^2(d)},
    $}
\end{equation}
where $t$ is the current number of training steps, $\mathrm{min}(d)$ represents the minimum value of $d$ in the training set, $T$ is the maximum number of training steps. 
At a given training step $t$, training examples with difficulty up to $c(t)$ (\ie, $d<c(t)$) will be sampled for training. As such, easier cases are more frequently fed to the model for training in the beginning. As instances with fewer columns in the schema are easier to derive schema dependencies with our algorithm, we expect the curriculum to help stabilize the pre-training process.

\section{Experiments}

\subsection{Datasets and Evaluation Metrics.}

We conduct experiments on two cross-domain table semantic parsing benchmarks, Spider and SQUALL. 
These benchmarks require generalizing to unseen tables/databases at dev and test time.
We adopt exact set match accuracy for Spider; logical form accuracy and execution accuracy for SQUALL.

\paragraph{Spider}
We conduct the experiments on the Spider \cite{yu2018spider} dataset, a large-scale benchmark for cross-domain complex text-to-SQL tasks. Spider consists of 11,840 examples which are split into training (size: 7,000), development (size: 2,134) and test set (size: 1,034), covering 138 different domains. In addition, SQL queries in the dataset are categorized into four difficulty levels based on the number of SQL keywords. Models are evaluated using the official exact matching accuracy metric of Spider. We conduct ablation studies on the development set since the test set is used for scoring models on the leaderboard and is not publicly accessible. 
We use SLSQL \citet{lei2020re} as the baseline model, which is a straightforward and robust approach exploring the role of schema linking in the Text-to-SQL model.
Meanwhile, we also compare our model with the state-of-the-art models on the Spider benchmark.

\paragraph{SQUALL}
SQUALL \cite{shi2020potential} is the first large-scale semantic parsing dataset with both hand-produced targets logical forms and manually derived lexical alignments between questions and SQL queries.
SQUALL consists of 15,622 examples which are split into training (size: 9,032), development (size: 2,246) and test set (size: 4,344).
For the baseline model, we apply SEQ2SEQ + BERT  mentioned as in \cite{shi2020potential} and adapt the same set of hyper-parameters, including batch size and learning rate.

\subsection{Implementation Details .} 
\label{impl}
Our model is initialized with 24-layer RoBERTa \citep{Liu2019}. We pre-train our model on 2 GPUs with each having a batch size of 6. 
Gradient clipping is applied to the model with a maximum gradient value of 1. To alleviate the overfitting issue, the maximum number of training steps is 500k. Moreover, a patient step number is set to 25k, i.e., if the metric does not increase for the pre-set step number, the training will carry out an early stop. We set the maximum learning rate to 1e-5. We also adopt a learning rate schedule that for the first 50k steps, the learning rate is linearly increased to the maximum one, then decayed for the left steps.

\subsection{Experiment Results and Analyses}
\paragraph{Overall Performance.}

From Table \ref{tab1}, \ref{tab:tab2}, we can observe that SDCUP substantially outperforms baseline models by a noticeable margin on exact match accuracy.
Enhanced by our SDCUP model, the base models can be improved over 3.5\% on Spider dataset and 1.7\% with the one with RoBERTa-large, and achieving the best results on both SQUALL and Spider, implying the overall benefits brought by MLM, SDP, SAC, and EPR. 
Specifically, we compare SDCUP with BERT-large, RoBERTa-large, GRAPPA~\cite{Yu2020} on SQUALL and Spider. 

\begin{table}[t]  
    \centering
    \small
    \footnotesize
    \scalebox{0.85}{
    \begin{tabular}{lc}  
    \toprule
    \textbf{Model}& \textbf{Dev.} \\ 
    \midrule
    EditSQL w/ BERT-base \citep{zhang2019editing} & 57.6 \\ 
    IRNet w/ BERT-base \citep{guo2019towards} & 61.9  \\ 
    RYANSQL w/ BERT-large \citep{Choi2020} & 70.6  \\
    TranX w/ TaBERT \citep{yin2020tabert} & 65.2  \\  
    RATSQL w/ BERT-base \citep{Wang2020} & 69.7\\
    \midrule
    SLSQL \cite{Wang2020}   \\
    \quad w/ BERT-large & 61.51  \\
    \quad w/ RoBERTa-large & 66.21  \\
    
    
    \quad w/ GRAPPA \cite{Yu2020} & 70.21  \\
    
    \quad w/ SDCUP (MLM) & 68.06 \\
    \quad w/ SDCUP (MLM+SDP) & 68.96 \\
    
    \quad w/ SDCUP (MLM+SDP+SAC) & 69.96\\
    
    \quad w/ SDCUP (MLM+SDP+SAC+EPR) & \textbf{70.70} \\
    \midrule
    \quad w/ Oracle & 72.40 \\
    \quad w/ SDCUP + Oracle & \textbf{75.92}\\
    \bottomrule
    \end{tabular}  
    }
    \caption{Exact match accuracy (\%) on Spider dev set. }
    \label{tab1}
\end{table}

Table \ref{tab1} shows the overall result on the Spider dataset.
Augmented with SDCUP, the model achieves significantly better performance than the baselines using BERT and RoBERTa, demonstrating its effectiveness on downstream tasks.
Our strongest model, SDCUP with (MLM+SDP+SAC+EPR), further enhances SLSQL with BERT-large by an absolute improvement of 9.2\% on the exact match accuracy on the dev set.
Even the PLM becomes larger (RoBERTa-large), SDCUP also outperforms it by 2.64\%.
Compared with the currently state-of-the-art pre-training method GRAPPA, SDCUP also obtains a competitive performance, outperform it by 0.5\%.
Taking into account that GRAPPA uses the dev-set data for pre-training, this result is awe-inspiring.

From the ~\cite{lei2020re}, we also report the result of SLSQL + BERT (Oracle), where the learnable schema linking module is replaced with human annotations in inference. 
It represents the maximum potential benefit of schema linking for the text-to-SQL task ~\cite{liu2021awakening}.
As mentioned, when introducing SDCUP, it dramatically improves the performance, up to almost 14.4\% by base model.
This surprising result demonstrates that SDCUP can fully activate the potential performance of SLSQL, thanks to the improvement of the perception of linking by large-scale pre-training.

Table \ref{tab:tab2} shows the results on SQUALL dataset.
By using large-scale pre-training models such as RoBERTa and SDCUP to encode question and schema, the performance of baseline can be boosted significantly (6\%). 
Compared with RoBERTa, our best model can further improve the performance by 1.8\%, and leading to a new state-of-the-art performance on this task. 
The experimental performance confirms the effectiveness of the object of SDP and EPR.

\begin{table}[t]
\centering
    \small
    \footnotesize
    \scalebox{0.85}{
\begin{tabular}{llll}
\toprule
\multirow{2}{*}{\textbf{Model}} & \multicolumn{2}{c}{\textbf{Dev.}} & \textbf{Test.} \\

                        & ACC$_{\text{LF}}$ &ACC$_{\text{EXE}}$  &ACC$_{\text{EXE}}$ \\
\midrule
  SEQ2SEQ               &  39.53         &  59.57        & 48.82 \\
  \quad w/ Align         &  43.70         &  62.10        & 50.10 \\
  \quad w/ BERT         &  45.99         &  64.78        & 54.05 \\
  \quad w/ (Align) BERT   &  48.40  &  67.70        & 54.30 \\
  \quad w/ RoBERTa      &  50.93         &  70.92        & 58.37 \\
  \quad w/ GRAPPA       &  52.76         &  \textbf{72.92}        & 59.32 \\
  \quad w/ SDCUP        & \textbf{52.95}  &71.12  & \textbf{59.56} \\

\bottomrule
\end{tabular}
}
\caption{Logical form (ACC$_{\text{LF}}$) and execution (ACC$_{\text{EXE}}$) accuracy (\%) on SQUALL dataset.  }
\label{tab:tab2}
\end{table}

\begin{table*}[ht]
\centering
\small
\begin{tabular}{p{2.1\columnwidth}} 
\toprule
\textbf{Question}: Show the stadium name and the number of concerts in each stadium. \\
\textbf{Basline}: SELECT stadium.Stadiu\_ID , count( $*$ ) FROM stadium JOIN concert  on concert.Stadium\_ID = stadium.Stadium\_ID  GROUP BY concert.Stadium\_ID   \\
\textbf{SDCUP}: SELECT stadium.Name , count( $*$ ) FROM stadium JOIN concert  on concert.Stadium\_ID = stadium.Stadium\_ID  GROUP BY stadium.Stadium\_ID     \\
\midrule

\textbf{Question}: Show the name and theme for all concerts and the number of singers in each concert.          \\
\textbf{Baseline}: SELECT concert.concert\_Name, concert.Theme, count( * ) FROM concert JOIN singer\_in\_concert  on singer\_in\_concert.concert\_ID = concert.concert\_ID  GROUP BY concert.concert\_Name          \\
\textbf{SDCUP}: SELECT concert.concert\_Name, concert.Theme, count( * ) FROM concert JOIN singer\_in\_concert  on singer\_in\_concert.concert\_ID = concert.concert\_ID  GROUP BY singer\_in\_concert.concert\_ID          \\
\midrule
\textbf{Question}: Find the weight of the youngest dog.          \\
\textbf{Baseline}: SELECT Pets.weight FROM Pets WHERE Pets.pet\_age = ( SELECT min ( Pets.pet\_age ) FROM Pets )          \\
\textbf{SDCUP}: SELECT Pets.weight FROM Pets ORDER BY Pets.pet\_age asc LIMIT 1          \\
\midrule
\textbf{Question}: What is the average age for all students who do not own any pets ?          \\
\textbf{Baseline}: SELECT avg ( Student.Age ) FROM Student WHERE Student.StuID $<$ ( SELECT avg ( Student.Age ) FROM Student )          \\
\textbf{SDCUP}: SELECT avg( Student.Age ) FROM Student WHERE Student.StuID not in ( SELECT Has\_Pet.StuID FROM Has\_Pet )          \\
\midrule
\textbf{Question}: Give the ids of documents that have between one and two paragraphs.         \\
\textbf{Baseline}: SELECT Paragraphs.Document FROM Paragraphs GROUP BY Paragraphs.Document\_ID HAVING count( * ) $>=$ 1          \\
\textbf{SDCUP}: SELECT Paragraphs.Document FROM Paragraphs GROUP BY Paragraphs.Document\_ID HAVING count( * ) between 1 and 2 \\
\bottomrule
    
\end{tabular}
 
\caption{Selected examples from Spider development set. }
\label{case}  
\end{table*}

\paragraph{Ablation Study.} 
We carry out ablation studies on our model. Concretely, we preserve MLM as the only objective to pre-train our model and fine-tune it on downstream tasks. SDCUP (MLM+SDP) outperforms SDCUP (MLM) with considerable margins on both Spider and SQUALL, suggesting that SDP efficiently helps the model capture structural bias between the question and schema and such bias is useful in pre-training for table semantic parsing tasks. Besides, SDCUP further enhanced by SAC can obtain much better performance on Spider and SQUALL, with the pre-training convergence time reduced by up to 30~\%.
With the help of EPR, SDCUP achieves the best performance on two datasets, demonstrating the importance of the correct order of entity.

\paragraph{Case Study.}

\begin{figure}[t]
    \centering
    \includegraphics[width=0.23\textwidth]{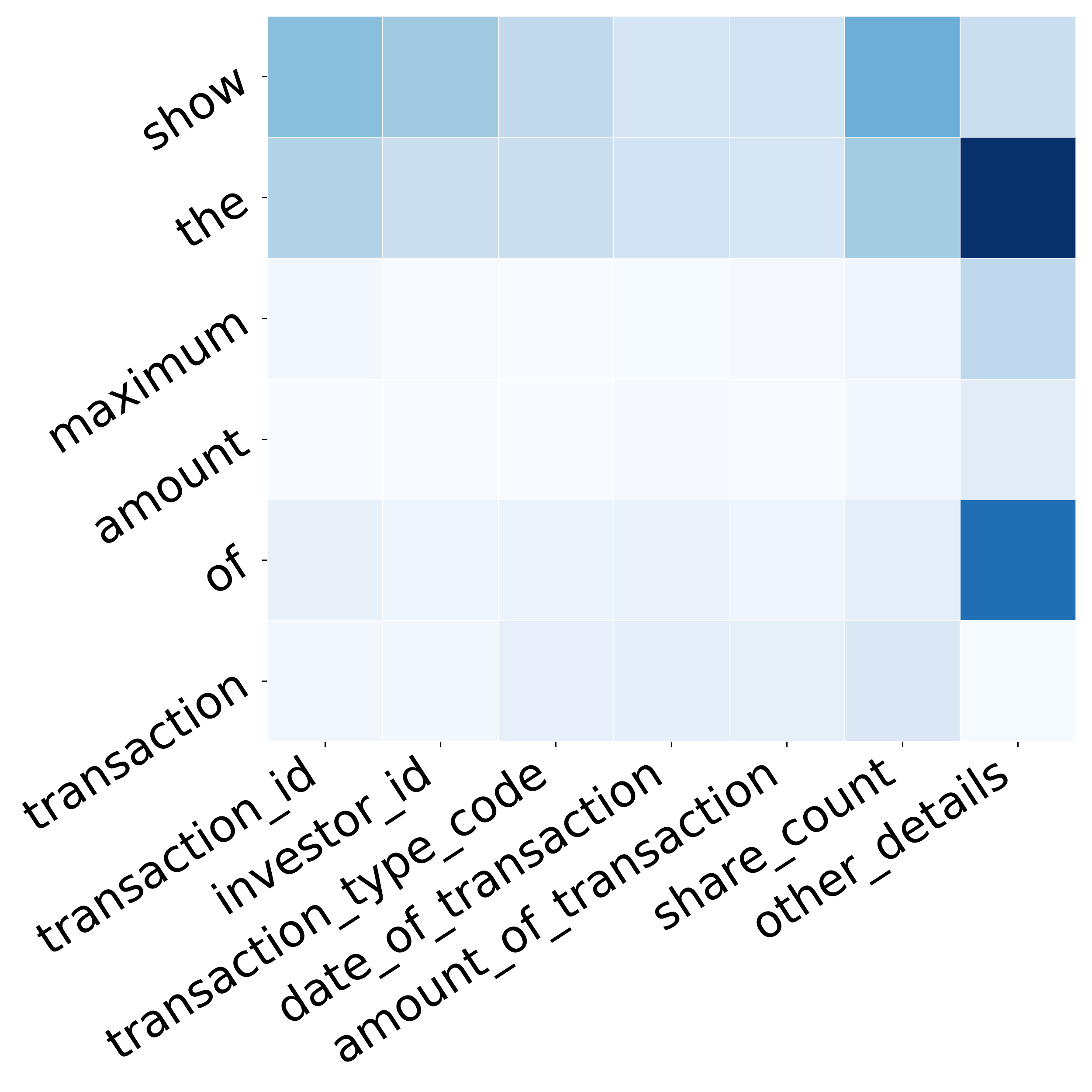}
    \includegraphics[width=0.23\textwidth]{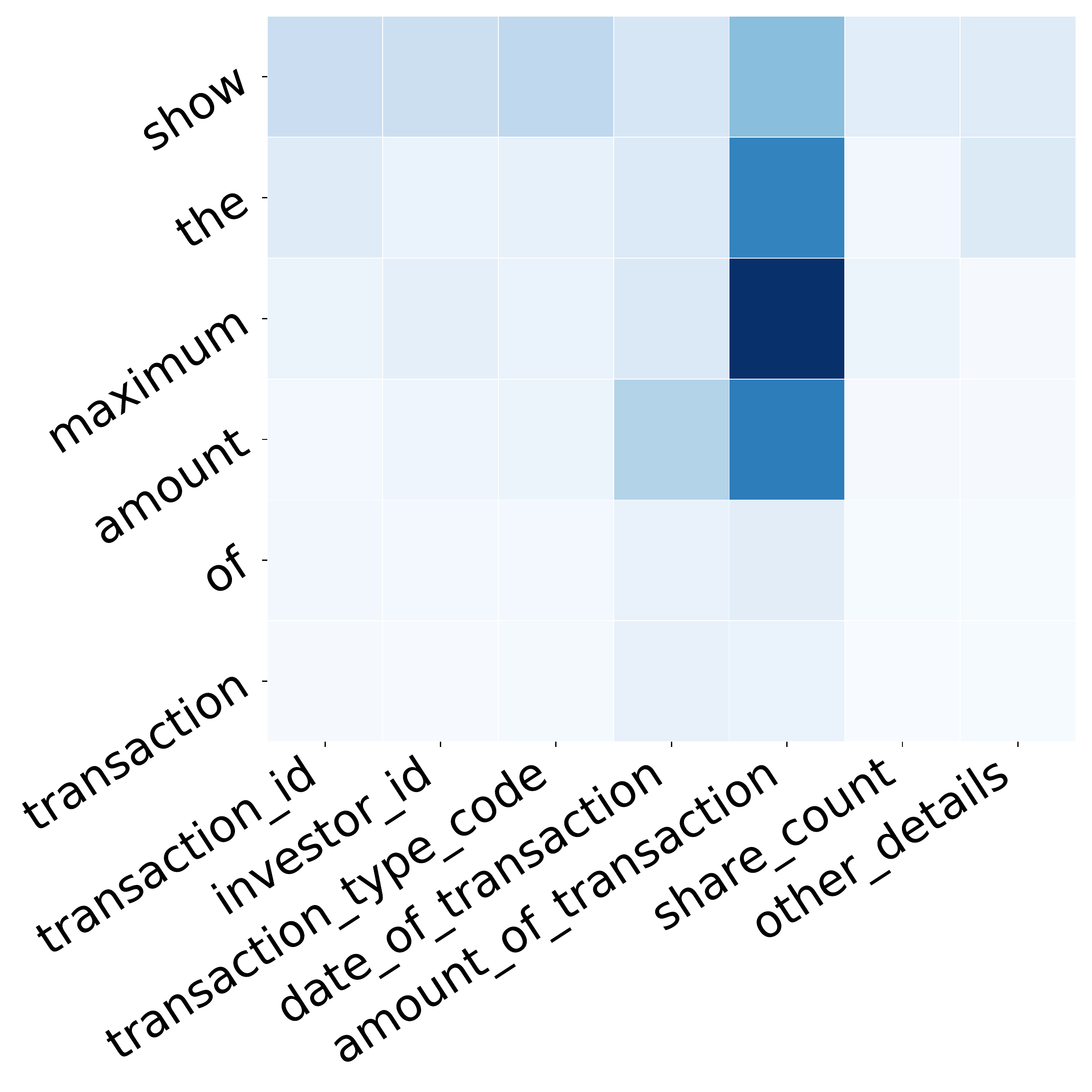}
    \caption{Attention visualization on the last self-attention layer. The left is from SDCUP (MLM) and the right is from SDCUP (MLM+SDP). Specifically, SDCUP (MLM) tends to ignore that the phrase \textit{the maximum amount of transaction} is highly related to \texttt{amount-of-transaction} while SDCUP (MLM+SDP) accurately aligns these question tokens with \texttt{amount-of-transaction}.    }
    \label{case_visual}
\end{figure}

To see whether attention distributions are influenced by the SDP objective,
We randomly select an instance from Spider dataset and visualize the attention interaction between the question and schema by truncating other details and normalizing the left. 
A key observation from Figure~\ref{case_visual} is that SDP largely regulates the learning of attention and biases it towards the distribution of schema dependency.
For the purpose of analyzing the effectiveness of our model, we further select some representative examples from the SPIDER development set, presented in Table~\ref{case}.
The baseline system refers to SLSQL+RoBERTa-large Encoder model, and our model refers to the SLSQL+SDCUP Encoder.
From the generated SQLs, we can conduct that our model achieves better schema linking performance, either explicit matching between the schema and question or implicit matching.
For example, in Example 1, the stadium name should match column stadium.Name instead of stadium.Stadiu\_ID, and in Example 2, the number of singers in each concert should match table singer\_in\_concert instead of concert.
Furthermore, our model can handle complex questions better, such as \texttt{ORDER BY}, \texttt{NOT IN}, and \texttt{BETWEEN AND} condition.

\section{Related Work}
Semantic parsing has long been a fundamental problem in NLP \citep{zelle1996learning,DBLP:conf/uai/ZettlemoyerC05,DBLP:conf/acl/WongM07,zettlemoyer2007online,berant2013semantic,li2014constructing,Yaghmazadeh2017SQLizerQS,iyer2017learning}. Recently, table semantic parsing (Text-to-SQL, \citealt{dong2018coarse,lin2020bridging}) has attracted extensive attention. 

To tackle the schema linking problem, \citet{guo2019towards} firstly used heuristic rules to construct intermediate representation.
\citet{bogin2019representing} and \citet{chen2021shadowgnn} introduced graph neural networks to model the structure of database schema.
\citet{lin2020bridging} leveraged the database content to augment the column representation.
The RATSQL \citep{Wang2020} designed explicit linking relations to handle various pre-defined relations. 
Furthermore, \citet{cao2021lgesql} proposed a line graph enhanced encoder to mine the underlying relational features by the heterogeneous graph.
\citet{lei2020re} re-examined the role of schema linking in Text-to-SQL model.
SeaD \citep{xuan2021sead} is the most relevant study to ours, who proposed schema-aware denoising sequence-to-sequence text-to-SQL generation.
Different from their work, we focus on the pre-training on the encoder side, which is more plausible to be applied on downstream tasks.
More recently, table semantic parsing in conversation scenarios has gradually emerged, where a series of context-dependent parsing methods were devised \citep{zhang2019editing,liu2020far,cai2020igsql,Wang2021TrackingIS,Hui2021DynamicHR}.

A new line of work is large-scale pre-training models.
Inspired by the fact that incorporating external knowledge can further improve the ability of pre-training models, \citet{xiong2019pretrained,Wang2020KAdapterIK,Rosset2020KnowledgeAwareLM} proposed table-based pre-training. 
\citet{yin2020tabert} and \citet{herzig2020tapas} introduced pre-trained encoders for a joint understanding of textual and tabular data. \citet{Shi2020} and \citet{Yu2020} proposed generation-augmented pre-training and grammar-augmented pre-training respectively.
\citet{deng2020structure} identified a column-value mapping objective to harness the text-table alignment knowledge.
\citet{wang2021learning} proposed a generative process for synthesizing question-SQL pairs and the synthesized data were further applied for data augmentation.
Distinguished from the above works, our work imposes schema dependency explicitly as an objective for pre-training to exploit structural bias and fill the gap between pre-training and downstream tasks.

\section{Conclusion}

In this paper, we show that the existing pre-training models for table semantic parsing can further benefit from modeling the structural and semantic relationships between questions and schemas. We propose schema dependency prediction and entity perturbation recovery as the main objective for pre-training. To circumvent instance-level noise associated with the data construction and dependency labeling procedure, we propose a schema-aware curriculum for pre-training. Experiments conducted on Spider and SQUALL exhibit the effectiveness of our proposed framework.

\bibliography{aaai22}
\end{document}